\documentclass{article} %
\usepackage{iclr2027_conference,times}

\usepackage{amsmath,amsfonts,bm}

\def\eqref#1{equation~\ref{#1}}

\def\1{\bm{1}}

\DeclareMathAlphabet{\mathsfit}{\encodingdefault}{\sfdefault}{m}{sl}
\SetMathAlphabet{\mathsfit}{bold}{\encodingdefault}{\sfdefault}{bx}{n}

\definecolor{cvprblue}{rgb}{0.21,0.49,0.74}
\definecolor{greenx}{RGB}{0,128,128}
\definecolor{maroonx}{RGB}{195,18,48}
\usepackage[colorlinks=True,
            linkcolor=maroonx,
            anchorcolor=blue,  
            pagebackref,
            citecolor=cvprblue,
            ]{hyperref}
\usepackage{url}
\usepackage{graphicx}
\usepackage{float}
\usepackage[dvipsnames]{xcolor}
\usepackage{booktabs}
\usepackage{changepage}

\usepackage{algorithm}
\usepackage{algpseudocode}

\usepackage{xspace}
\usepackage{listings}
\usepackage{enumitem}
\setlist[itemize]{leftmargin=*,topsep=0em}
\usepackage{amssymb}
\usepackage{colortbl}
\usepackage{multirow}
\usepackage{xcolor}
\definecolor{skillcreate}{HTML}{F66B32}
\definecolor{skillreuse}{HTML}{08AF82}
\usepackage{colortbl}
\usepackage{wrapfig}
\usepackage{makecell}

\definecolor{rankblueone}{HTML}{ACD4FF}
\definecolor{rankbluetwo}{HTML}{CAE4FF}
\definecolor{rankbluethree}{HTML}{E8F3FF}
\definecolor{sftpink}{HTML}{FBE1E7}
\newcommand{\sftbest}[1]{\cellcolor{sftpink}#1}

\newcommand{\rankfirst}[1]{\cellcolor{rankblueone}#1}
\newcommand{\ranksecond}[1]{\cellcolor{rankbluetwo}#1}
\newcommand{\rankthird}[1]{\cellcolor{rankbluethree}#1}

\usepackage{titlesec}
\titlespacing{\section}{0pt}{4pt}{3pt}
\titlespacing{\subsection}{0pt}{2pt}{1pt}

\newcommand{\titleicon}{\raisebox{-0.18\height}{\includegraphics[height=1.35em]{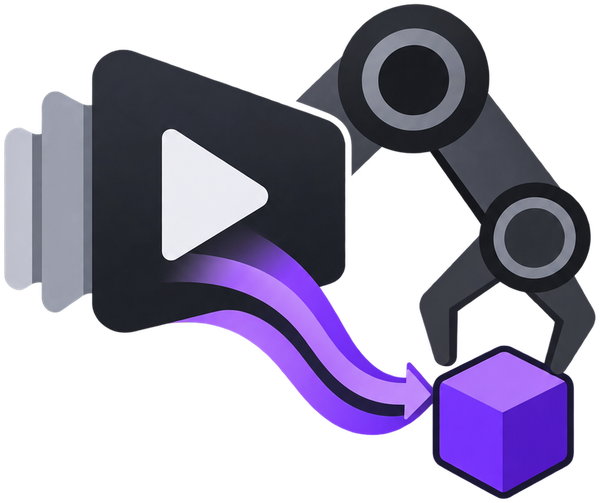}}\hspace{0.35em}}
\title{\titleicon\texttt{Video2Skill}: From Streaming Experience to Reusable Embodied Skills}

\author{%
Jianshu Zhang\textsuperscript{1}$^{*}$\quad
Ce Zhang\textsuperscript{2}$^{*}$\quad
Xiyuan Yang\textsuperscript{3}\quad
Chenwei Xu\textsuperscript{1}\quad
Haoran Lu\textsuperscript{1}\quad
Yijiang Li\textsuperscript{4} \\
\bfseries Yaqi Xie\textsuperscript{2}\quad
Katia P.~Sycara\textsuperscript{2}\quad
Han Liu\textsuperscript{1} \\
\normalfont \textsuperscript{1}Northwestern\qquad
\textsuperscript{2}CMU\qquad
\textsuperscript{3}UIUC\qquad
\textsuperscript{4}UCSD\qquad
$^{*}$Equal contribution \\
\normalfont\fontfamily{pcr}\selectfont Project Page: \href{https://andyzworks.github.io/video2skill/}{\textcolor{red}{https://andyzworks.github.io/video2skill}}%
}

\iclrfinalcopy
\begin{document}

\maketitle
\vspace{-24pt}

\begin{abstract}
Manipulation behaviors vary widely across objects and scenes, but they share a small set of reusable skills, and planning with these skills helps embodied agents generalize to new tasks. Yet an agent can only plan with skills it knows. Recovering skills from observed experience, the inverse of planning, builds this knowledge over time and yields skill data for training future agents. Vision-Language Models (VLMs) describe individual manipulation events well, but can they organize a stream of events into reusable skills? We formulate this problem as \emph{Streaming Embodied Skill Discovery (SESD)}: a model watches videos in sequence and maintains a persistent skill library that shapes its later decisions. To systematically measure this ability, we introduce \texttt{Video2Skill}, a benchmark that covers robot tabletop manipulation and human kitchen activity and tests three core capabilities: (i) locating manipulation events in time, (ii) grouping events of the same transformation, and (iii) deciding when to reuse an existing skill or create a new one. Across 19 open-source VLMs, many models group events at near-chance level, and scale does not consistently help. Their errors depend on how perception and library updates are coupled: joint models merge distinct transformations into one skill, while models that update the library from text descriptions duplicate recurring ones. Supervised fine-tuning, including our counterfactual library-state rebalancing (\textsc{CLaRe}), improves grouping but exposes a deeper bottleneck: trained models consolidate familiar skills yet rarely expand the library. Their libraries stall below half the reference size, and transformations unseen in training are located in time but almost never given a new skill. Recognizing when existing skills are insufficient thus emerges as the central challenge. \looseness=-1
\end{abstract}

\begin{figure}[H]
\vspace{-12pt}
    \centering
    \includegraphics[width=\linewidth]{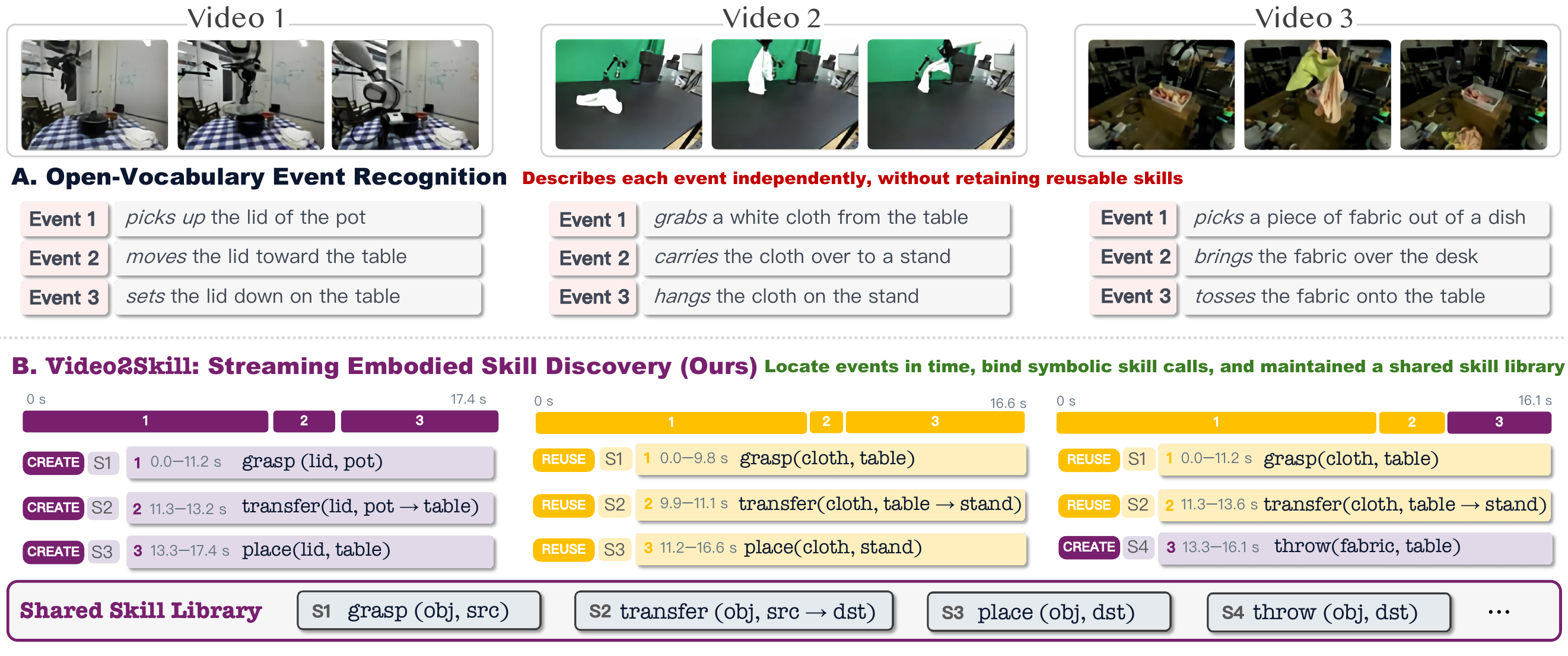}
    \vspace{-18pt}
    \caption{\textbf{\texttt{Video2Skill}: From describing events to accumulating reusable skills.}
    (A) Vision-Language Models recognize manipulation events in open-vocabulary language.
    (B) \texttt{Video2Skill} further organizes these observations into a persistent library of symbolic skills, grounding each event in time and deciding whether to reuse an existing skill or introduce a new one.
    Video~1 introduces \texttt{grasp}, \texttt{transfer}, and \texttt{place}; Video~2 reuses all three despite changes in objects and scene; Video~3 reuses \texttt{grasp} and \texttt{transfer} while adding \texttt{throw} for a new transformation.
    The library thus preserves shared structure across observations while expanding to accommodate novelty.
    }
    \label{fig:teaser}
\end{figure}

\clearpage
\section{Introduction}
Embodied agents that carry out long-horizon robotic tasks must plan. The behaviors they encounter are enormously diverse, varying in objects, scenes, and embodiments, yet they are underpinned by a much smaller set of shared skills: wiping a table, a plate, or a window all instantiate the same wiping skill. Planning over such skills therefore generalizes, as new tasks can be solved by recomposing familiar skills rather than learning each behavior from scratch~\citep{ichter2023can,liang2023code}. Planning with skills, however, presupposes knowing which skills exist and how they are instantiated. A natural precursor is the inverse problem: observing continuous experience and recovering the skills that produced it~\citep{baker2009action}. Much as people acquire skills by watching others and reuse them in new settings, an agent that solves this inverse problem can accumulate a skill library as experience unfolds, and the resulting skill-annotated experience can supervise future embodied agents~\citep{kim2025uniskill,xie2026uniskill}. These skills are most useful when represented symbolically and with structure: a schema such as \texttt{grasp(object, source)} can be recognized across instances, composed by planners, and inspected by humans~\citep{yang2025skillwrapper,yu2026memora}. \looseness=-1

Vision-Language Models (VLMs) can already describe diverse manipulation events in open-vocabulary language~\citep{bai2025qwen2,chen2024internvl,zhang2025video,pi2024image}. For agents operating over continuous visual streams~\citep{qian2025dispider,wang2026from,wang2026egomemreason,zhang2026streamscout,zhang2026vscan}, a further challenge is to organize these events into a persistent skill library. Doing so requires locating events in time and deciding whether each instantiates a known skill or introduces a new one. Figure~\ref{fig:teaser} illustrates this distinction: picking up a lid and grasping a cloth should reuse the same \texttt{grasp} schema with different arguments, whereas throwing fabric onto a table and placing it there involve distinct motion and release processes despite sharing a destination. These decisions become more challenging in a streaming setting, where the model relies on a library built from its own earlier predictions. Creating duplicate schemas fragments recurring experience, while reusing an overly broad schema merges distinct transformations. Because subsequent decisions depend on this evolving library, such errors can shape how later events are interpreted. %

Existing approaches to skill reuse often rely on predefined repertoires or on signals beyond passive video observations. Some systems compose an existing set of robot skills~\citep{ichter2023can,liang2023code}, while others grow their repertoires through code execution and environment feedback~\citep{wang2024voyager,ning2026code}, task-success signals~\citep{zhang2023bootstrap}, or robot trajectories~\citep{wan2024lotus}. These settings differ from constructing a symbolic skill library online from a stream of visual observations. This leaves a fundamental question: \textit{can a model build and maintain such a library from streaming video, using the abstractions established by its own earlier decisions to recognize familiar transformations and determine when a new schema is needed?}

We formulate this problem as \emph{Streaming Embodied Skill Discovery (SESD)}. A model incrementally processes video, predicts the temporal extent of each manipulation event, and expresses it as a symbolic skill call: a reusable transformation schema bound to concrete arguments such as the object, source, and destination. The library persists across videos and conditions subsequent predictions, allowing schemas discovered in one scene to be reused in another. Skills here are semantic abstractions of observed transformations, rather than executable controllers. We introduce \texttt{Video2Skill}, a benchmark built from robot manipulation videos in \textsc{RoboInter} and egocentric kitchen recordings in \textsc{HD-EPIC}, to evaluate this capability across robotic and human activity. The benchmark evaluates how well models recover manipulation events, group instances of the same transformation regardless of schema names, and distinguish when to create or reuse a skill.

We evaluate 19 open-source VLMs using two processing paradigms: \emph{unified} processing jointly recognizes events and updates the library, whereas \emph{factorized} processing first generates textual event descriptions without library access and then uses the same model to update the library from those descriptions. Without task-specific training, many state-of-the-art models exhibit near-chance agreement with the reference grouping of events, revealing a substantial gap between event description and consistent skill abstraction. The two paradigms exhibit contrasting failure tendencies: unified models tend to merge distinct transformations into overly broad skills, while factorized models frequently create duplicate schemas for recurring transformations.\looseness=-1

To investigate whether supervision can mitigate these failures, we fine-tune Qwen3.5-4B and LLaVA-OneVision-2-8B under both paradigms using three strategies: oracle-history supervised fine-tuning (SFT), Counterfactual Library-state Rebalancing (\textsc{CLaRe}), and on-policy correction using student-generated histories. Supervision improves skill grouping across both backbones, but models still struggle to distinguish new transformations from variations of familiar ones. The benefits of the three strategies vary across settings, highlighting the difficulty of learning when to preserve an existing skill and when to expand the library. Further analyses reveal where these difficulties persist. Training makes the schemas produced under different video orders more consistent, but its benefits are uneven across skills, with the most consistent gains on frequently demonstrated transformations. Transformations absent from task-specific training are often absorbed into existing schemas, even when their events are located in the video. These findings suggest that models can become better at organizing experience without reliably recognizing the limits of their existing repertoire. Developing models that both consolidate recurring transformations and introduce appropriate new abstractions remains a central challenge for learning from streaming visual experience.

Our contributions can be summarized as follows:
\begin{itemize}[leftmargin=*,itemsep=1pt,topsep=2pt]
\item We formulate \emph{Streaming Embodied Skill Discovery} (SESD) and introduce \texttt{Video2Skill}, a benchmark built from \textsc{RoboInter} and \textsc{HD-EPIC}. Its evaluation jointly measures event coverage, naming-invariant skill grouping, and creation--reuse decisions across streaming videos.
\item We evaluate 19 open-source VLMs under unified and factorized paradigms, identifying contrasting failure tendencies: unified models often merge distinct transformations, while factorized models frequently assign recurring transformations to duplicate schemas.
\item We compare three supervision strategies across two backbones and both paradigms, finding improved grouping alongside persistent creation--reuse errors. Further analyses examine order consistency, uneven gains across training frequencies, and limited novelty recognition for transformations absent from task-specific supervision.
\end{itemize}

\section{Problem Definition and Benchmark Curation}
\label{sec:benchmark}
We introduce \texttt{Video2Skill} to study whether models can organize streaming visual experience into reusable skills. We first formalize Streaming Embodied Skill Discovery (SESD), then describe its construction from human and robot manipulation videos, and finally present an evaluation protocol covering event recovery, skill grouping, and creation--reuse decisions.

\subsection{Task Formulation}
\label{sec:task}

We process an ordered video stream $\mathcal{V}=(v_1,\ldots,v_N)$ in chunks. At step $t$, the model receives chunk $x_t$ and state $z_{t-1}=(\mathcal{L}_{t-1},u_{t-1})$, containing a persistent skill library and an optional unfinished event:
\begin{equation}
y_t\sim\pi_\theta(\cdot\mid x_t,z_{t-1}),
\qquad
y_t=(\mathcal{D}_t,\mathcal{E}_t,u_t).
\end{equation}
Here, $\mathcal{D}_t$ contains new schemas, $\mathcal{E}_t$ contains completed events, and $u_t$ carries an unfinished event across chunks. Each event $e=(b_e,d_e,s_e,a_e)$ specifies video-relative temporal boundaries, a schema, and argument bindings.

A schema $s=(n_s,d_s,\mathcal{A}_s)$ comprises a stable free-form name, a definition, and typed argument slots. It describes a physical transformation, not an executable controller. Reference equivalence is $e_i\sim e_j \iff \tau(e_i)=\tau(e_j)$, where $\tau(e)$ denotes the underlying transformation, independently of the model's schema assignments.

Starting from $\mathcal{L}_0=\varnothing$ and $u_0=\varnothing$, the state updates as
\begin{equation}
\mathcal{L}_t=\mathcal{L}_{t-1}\cup\mathcal{D}_t,
\qquad z_t=(\mathcal{L}_t,u_t).
\label{eq:update}
\end{equation}
Events reference schemas in $\mathcal{L}_t$, either newly created or reused. At video boundaries, unfinished events are closed or discarded and $u_t$ is reset; the library persists. Thus, earlier creation and reuse decisions shape the context for subsequent observations.

\subsection{Benchmark Curation}
\label{sec:curation}

\textbf{Source datasets.} We use two public datasets spanning human and robot manipulation: \textsc{HD-EPIC}~\citep{perrett2025hd}, with long egocentric recordings of unscripted kitchen activity, and \textsc{RoboInter}, with robot tabletop videos drawn from DROID and RH20T~\citep{khazatsky2024droid,fang2024rh20t}. The domains differ in action density and duration: median reference segments in the verified test splits last 1.9\,s and 4.5\,s, respectively.

\textbf{Data curation.} Our annotation process prioritizes visual evidence over preserving source labels. For \textsc{HD-EPIC}, we annotate selected recordings and clips throughout. For \textsc{RoboInter}, we check each candidate source segment against the video, correct annotation errors, and discard labels whose stated manipulation is not visually supported. Retained events are represented as typed skill calls, with argument bindings drawn from seven roles: object, source, destination, instrument, direction, substance, and state. Review corrected 241 of the 12{,}429 verified calls across the training and test splits. The retained \textsc{RoboInter} segments cover 55\% of training video time and 82\% of verified test video time, in contrast to the wall-to-wall native annotations. These references therefore represent verified manipulations rather than exhaustive temporal coverage. We organize verified calls into 48 canonical transformation classes with definitions and aliases, establishing the reference partition for naming-invariant evaluation. Models generate free-form schemas without receiving this class list.

\textbf{Data splits.} \textsc{HD-EPIC} provides 42 fully annotated training videos with 5{,}726 calls and 40 test clips from 29 held-out recordings with 1{,}137 reference segments. \textsc{RoboInter} provides 2{,}992 training videos with 5{,}032 verified calls and 132 test videos with 534 verified segments. The verified test splits contain 46 canonical classes in \textsc{HD-EPIC} and 15 in \textsc{RoboInter}. All training and test splits are disjoint at the source-video level, as verified programmatically.

\subsection{Evaluation Protocol}
\label{sec:evaluation}
We evaluate SESD along three complementary dimensions: event coverage, naming-invariant skill grouping, and creation--reuse decisions.

\textbf{Temporal alignment and coverage.}
For video $v$, let $\mathcal{R}_v$ and $\mathcal{P}_v$ denote reference and predicted segments. Each reference independently selects the maximum-tIoU prediction, accepting matches at IoU $\geq 0.3$. Let $\mathcal{M}$ contain matched references across the episode:
\begin{equation}
p^*(r)=\arg\max_{p\in\mathcal{P}_v}\operatorname{tIoU}(r,p),
\qquad
\mathrm{Coverage}=\frac{|\mathcal{M}|}{\sum_v|\mathcal{R}_v|}.
\end{equation}
Predictions may match multiple references. Unmatched references enter only the coverage denominator; unmatched predictions incur no penalty. Coverage measures reference recovery.

\textbf{Skill grouping.}
Index the $m=|\mathcal{M}|$ matched events in episode order, with reference classes $c_i$ and aligned predicted names $\hat{s}_i$. Define
\begin{equation}
\begin{aligned}
\mathcal{Q}_{\mathrm{ref}}&=\{(i,j):1\leq i<j\leq m,\ c_i=c_j\},\\
\mathcal{Q}_{\mathrm{pred}}&=\{(i,j):1\leq i<j\leq m,\ \hat{s}_i=\hat{s}_j\}.
\end{aligned}
\end{equation}
Including cross-video pairs, we compute
\begin{equation}
P_{\mathrm{pair}}=
\frac{|\mathcal{Q}_{\mathrm{ref}}\cap\mathcal{Q}_{\mathrm{pred}}|}
{|\mathcal{Q}_{\mathrm{pred}}|},
\qquad
R_{\mathrm{pair}}=
\frac{|\mathcal{Q}_{\mathrm{ref}}\cap\mathcal{Q}_{\mathrm{pred}}|}
{|\mathcal{Q}_{\mathrm{ref}}|}.
\end{equation}
Low precision indicates overmerging; low recall indicates fragmentation. Adjusted Rand Index (ARI) corrects partition agreement for chance under fixed cluster sizes. Grouping is invariant to consistent renaming, but synonymous strings remain distinct. Pair counts scale as $\binom{n}{2}$, favoring frequent classes; ARI does not ensure class balance.

\textbf{Creation and reuse decisions.} We traverse matched events in episode order and independently identify first occurrences of reference classes and predicted names:
\begin{equation}
g_i=\mathbf{1}[c_i\notin\{c_j:j<i\}],
\qquad
\hat{g}_i=\mathbf{1}[\hat{s}_i\notin\{\hat{s}_j:j<i\}].
\end{equation}
One denotes \textsc{Create} and zero denotes \textsc{Reuse}. We report \textsc{Create} recall, $\Pr(\hat{g}_i=1\mid g_i=1)$; \textsc{Reuse} recall, $\Pr(\hat{g}_i=0\mid g_i=0)$; and decision accuracy, the fraction with $\hat{g}_i=g_i$. These measure novelty within the matched sequence, not explicit operations or complete library history, and exclude \textsc{Continue} evaluation.

\begin{table*}[t]
\centering
\setlength{\fboxsep}{2pt}
\caption{\textbf{Raw model performance on SESD.}
Unified models jointly identify events and update the skill library;
factorized models first propose events without library access, then use
the same model for text-only library updates.
Results use verified test splits in balanced order.
Create R and Reuse R denote recall.
All scores are multiplied by 100; ARI can be negative.
Within each paradigm and column,
\protect\colorbox{rankblueone}{first},
\protect\colorbox{rankbluetwo}{second}, and
\protect\colorbox{rankbluethree}{third} ranks are highlighted.}
\vspace{5pt}
\small
\setlength{\tabcolsep}{4pt}
\renewcommand{\arraystretch}{1.02}
\resizebox{\textwidth}{!}{%
\begin{tabular}{lcccccccccccc}
\toprule
& \multicolumn{6}{c}{\textsc{RoboInter}}
& \multicolumn{6}{c}{\textsc{HD-EPIC}} \\
\cmidrule(lr){2-7}\cmidrule(lr){8-13}
Model
& Cov. & Pair P & Pair R & ARI & Create R & Reuse R
& Cov. & Pair P & Pair R & ARI & Create R & Reuse R \\
\midrule
\rowcolor{gray!10}
\multicolumn{13}{l}{\textit{\textbf{Unified}: a single VLM watches the video and manages the library in one pass}} \\
Qwen3.5-0.8B
& 2.4 & 36.4 & \rankthird{77.4} & -10.0 & \ranksecond{25.0} & 88.9
& 0.2 & 0.0 & 0.0 & 0.0 & \rankfirst{50.0} & 0.0 \\
Qwen3.5-2B
& 39.3 & 31.1 & 72.8 & 8.0 & 7.7 & \rankthird{99.5}
& 14.8 & 9.2 & 53.8 & 5.1 & 9.1 & \rankfirst{99.3} \\
Qwen3.5-4B
& \rankthird{56.9} & \rankthird{36.7} & 63.1 & \ranksecond{19.4} & 20.0 & 99.3
& \rankfirst{61.4} & 12.5 & 32.7 & 8.7 & 11.1 & \rankthird{98.5} \\
Qwen3.5-9B
& 44.8 & 24.9 & 34.2 & 1.7 & \rankthird{21.4} & 99.1
& 53.9 & 15.5 & 26.1 & 11.8 & 20.0 & 98.2 \\
Qwen3.5-27B
& 47.9 & 30.9 & 19.0 & 4.6 & \rankthird{21.4} & 93.8
& 51.3 & \ranksecond{26.9} & 22.9 & \rankthird{19.9} & \ranksecond{37.2} & 95.6 \\
Qwen3.5-35B-A3B
& \ranksecond{57.1} & 29.6 & 37.1 & 4.3 & 13.3 & 99.0
& \rankthird{55.9} & 19.7 & 19.9 & 13.9 & 29.5 & 96.1 \\
Qwen3-VL-8B
& 56.0 & 30.0 & 59.1 & 12.4 & 20.0 & \rankfirst{100.0}
& 44.1 & 13.1 & 48.9 & 11.5 & 12.2 & 97.2 \\
Qwen3.8-27B
& \rankfirst{63.7} & \rankfirst{44.6} & 34.2 & \rankthird{18.7} & 15.4 & 96.9
& \ranksecond{60.1} & \rankfirst{30.0} & 29.6 & \ranksecond{24.3} & \ranksecond{37.2} & 94.7 \\
InternVL3.5-4B
& 41.6 & 36.0 & 72.1 & 12.6 & 15.4 & \rankthird{99.5}
& 19.0 & 7.6 & 54.7 & 2.3 & 16.2 & 98.3 \\
InternVL3.5-8B
& 48.5 & 35.1 & 62.8 & 11.0 & 13.3 & \rankfirst{100.0}
& 26.6 & 6.9 & 42.0 & 0.5 & 8.6 & 97.4 \\
InternVL3.5-14B
& 50.9 & 28.8 & 43.3 & 6.1 & 16.7 & 98.5
& 38.0 & 12.5 & 33.6 & 9.0 & 17.5 & 96.9 \\
InternVL3.5-38B
& 41.4 & 35.9 & 74.7 & 7.2 & 15.4 & 97.1
& 42.3 & 15.8 & 24.1 & 11.1 & 16.3 & 96.8 \\
Ovis2.5-9B
& 4.9 & 24.7 & 45.1 & -1.1 & 20.0 & 90.5
& 36.5 & 6.4 & \rankfirst{66.9} & -0.5 & 4.5 & \ranksecond{98.9} \\
Cosmos-Reason2-8B
& 48.5 & \ranksecond{43.2} & \ranksecond{80.3} & \rankfirst{27.0} & 6.7 & \ranksecond{99.6}
& 37.6 & 8.7 & \rankthird{58.2} & 4.4 & 20.5 & 98.2 \\
Cosmos-Reason2-32B
& 54.3 & 32.5 & 34.6 & 2.9 & \rankthird{21.4} & 96.7
& 44.4 & 12.0 & 31.0 & 8.5 & 24.4 & 92.5 \\
LLaVA-OneVision-2-8B
& 37.8 & 22.5 & \rankfirst{100.0} & 0.0 & 7.1 & \rankfirst{100.0}
& 21.6 & 13.4 & 52.4 & 13.2 & 12.8 & 95.7 \\
GLM-4.1V-9B
& 40.1 & 28.1 & 76.8 & 5.1 & 7.1 & 97.5
& 5.8 & 4.2 & 21.2 & -0.7 & \rankthird{36.4} & 84.1 \\
MiniCPM-V-4.5
& 43.8 & 25.4 & 58.6 & 7.5 & \rankfirst{26.7} & \rankthird{99.5}
& 7.5 & \rankthird{24.7} & \ranksecond{60.6} & \rankfirst{25.6} & 26.1 & 93.5 \\
Gemma-4-31B
& 52.6 & 36.6 & 36.6 & 12.5 & 7.1 & 95.9
& 46.0 & 15.5 & 25.6 & 12.0 & 34.9 & 94.0 \\
\midrule
\rowcolor{gray!10}
\multicolumn{13}{l}{\textit{\textbf{Factorized}: the VLM first proposes events then updates the library via a text-only meta-policy}} \\
Qwen3.5-0.8B
& 26.0 & 0.0 & 0.0 & -0.4 & \rankthird{75.0} & 13.4
& 11.1 & \rankfirst{50.0} & 0.6 & 0.9 & \rankfirst{93.8} & 6.4 \\
Qwen3.5-2B
& 42.3 & 23.1 & 0.2 & -0.2 & 71.4 & 15.6
& 32.4 & \ranksecond{48.0} & 2.4 & 4.0 & \ranksecond{92.1} & 18.2 \\
Qwen3.5-4B
& 46.6 & 31.3 & 8.9 & 3.2 & 26.7 & 68.8
& 50.1 & 6.3 & \rankthird{15.9} & -1.0 & 44.2 & 47.6 \\
Qwen3.5-9B
& 50.2 & 24.3 & 10.1 & 0.5 & 46.7 & 70.8
& 49.2 & 21.4 & 4.3 & 5.0 & 62.8 & 41.5 \\
Qwen3.5-27B
& 47.2 & 31.3 & 15.6 & 3.9 & 35.7 & \ranksecond{90.3}
& \rankthird{52.0} & 30.9 & \ranksecond{16.1} & \ranksecond{17.3} & 60.5 & \ranksecond{80.3} \\
Qwen3.5-35B-A3B
& 57.3 & 23.1 & 0.4 & -0.1 & 60.0 & 36.8
& \rankfirst{57.8} & 34.4 & 2.0 & 3.1 & 79.5 & 37.0 \\
Qwen3-VL-8B
& 52.2 & 32.2 & \ranksecond{18.2} & \ranksecond{6.9} & 33.3 & 82.6
& 49.1 & 23.9 & 12.3 & 12.5 & 62.5 & 66.4 \\
Qwen3.8-27B
& 43.5 & 28.9 & 16.7 & 0.8 & 33.3 & \ranksecond{90.3}
& \ranksecond{52.7} & 29.7 & \rankfirst{22.3} & \rankfirst{20.8} & 40.9 & \rankfirst{87.6} \\
InternVL3.5-4B
& 41.9 & \ranksecond{51.4} & 15.6 & \rankfirst{7.9} & 46.2 & 70.6
& 33.5 & 32.9 & 5.7 & 8.0 & 71.8 & 28.9 \\
InternVL3.5-8B
& 38.4 & \rankfirst{53.7} & 10.5 & 5.5 & 50.0 & 51.8
& 34.7 & 25.3 & 3.3 & 4.4 & 76.9 & 27.8 \\
InternVL3.5-14B
& 47.9 & 36.3 & \rankthird{17.9} & 4.6 & 42.9 & \rankthird{86.0}
& 51.2 & 26.3 & 2.8 & 3.7 & 76.2 & 38.9 \\
InternVL3.5-38B
& 48.3 & \rankthird{44.9} & 7.5 & 4.1 & 61.5 & 62.9
& 49.7 & 27.7 & 5.1 & 6.5 & 67.4 & 45.8 \\
Ovis2.5-9B
& \rankthird{65.2} & 41.4 & 8.0 & 5.1 & 50.0 & 47.3
& 47.5 & \rankthird{35.3} & 1.9 & 3.0 & 74.4 & 23.9 \\
Cosmos-Reason2-8B
& 52.8 & 31.9 & 0.3 & -0.0 & 64.3 & 26.5
& 50.7 & 21.3 & 4.9 & 5.6 & 63.6 & 44.0 \\
Cosmos-Reason2-32B
& 47.2 & 40.6 & 4.1 & 1.7 & 53.8 & 67.4
& 28.8 & 17.5 & 6.5 & 6.2 & 71.0 & 45.3 \\
LLaVA-OneVision-2-8B
& \rankfirst{71.7} & 27.4 & 0.8 & 0.1 & \ranksecond{80.0} & 29.1
& 40.5 & 17.2 & 0.6 & 0.7 & 87.8 & 16.0 \\
GLM-4.1V-9B
& 36.5 & 17.4 & 0.1 & -0.1 & \rankfirst{85.7} & 11.6
& 25.2 & 19.2 & 0.2 & 0.3 & 85.4 & 7.7 \\
MiniCPM-V-4.5
& \ranksecond{67.4} & 30.1 & 0.2 & 0.1 & \ranksecond{80.0} & 14.8
& 43.0 & 26.3 & 0.2 & 0.4 & \rankthird{88.1} & 9.2 \\
Gemma-4-31B
& 54.9 & 28.4 & \rankfirst{21.6} & \rankthird{5.8} & 35.7 & \rankfirst{93.9}
& 51.9 & 22.5 & 15.3 & \rankthird{13.8} & 50.0 & \rankthird{75.5} \\
\bottomrule
\end{tabular}%
}
\vspace{-15pt}
\label{tab:bigsweep}
\end{table*}

\section{How Do Current Models Behave?}
\label{sec:current}

We evaluate 19 off-the-shelf VLMs on both domains using two inference paradigms (Table~\ref{tab:bigsweep}). \emph{Unified} models jointly ground events and update the library from visual input and persistent state; \emph{factorized} models first propose events without library access, then use the same model for text-only library updates. All evaluations use verified test splits, balanced video order, and no task-specific training. We examine pair precision and recall separately to distinguish overmerging from fragmentation.

\textbf{Unified models tend to overmerge; factorized models tend to fragment.}
Unified models often reuse existing schemas even when a new transformation appears. Within the matched sequence, this tendency produces high reuse recall but low creation recall, while low pair precision confirms that distinct transformations are frequently grouped together. Factorized models create new schemas more readily, often improving pair precision. However, reuse and pair recall decline in nearly every comparison: the models become more willing to distinguish new transformations, but also repeatedly create separate schemas for transformations they have already encountered.

For example, on \textsc{HD-EPIC}, switching Qwen3.5-2B from unified to factorized raises pair precision from 9.2 to 48.0 but reduces pair recall from 53.8 to 2.4. Creation recall rises from 9.1 to 92.1, while reuse recall falls from 99.3 to 18.2. Higher creation recall thus accompanies a failure to reuse schemas for recurring transformations: the model separates different transformations more precisely, but also splits repeated instances across different names. 

\textbf{Scaling does not consistently improve skill abstraction.}
For unified Qwen3.5 on \textsc{HD-EPIC}, scaling from 4B to 27B increases pair precision from 12.5 to 26.9 but reduces pair recall from 32.7 to 22.9. Among matched events, the larger model exhibits less overmerging but greater fragmentation. ARI improves from 8.7 to 19.9, whereas the same scaling comparison on \textsc{RoboInter} reduces ARI from 19.4 to 4.6. InternVL3.5 likewise shows no monotonic improvement in ARI with size. Larger models can improve individual metrics, but do not consistently recover better skill partitions; ARI remains at or below 27.0 across all configurations.

\textbf{Coverage and grouping quality must be interpreted together.}
Factorized LLaVA-OneVision-2-8B achieves the highest coverage on \textsc{RoboInter} (71.7\%), yet its pair recall is only 0.8 and ARI is 0.1. Its unified counterpart exhibits the opposite problem: pair recall reaches 100.0, but pair precision is 22.5 and ARI is 0.0. Recovering more events therefore does not ensure consistent grouping, while high pair recall can result from grouping distinct transformations together.
Partition scores also depend on which events are covered. Unified MiniCPM-V-4.5 achieves the highest ARI on \textsc{HD-EPIC} (25.6) while covering only 7.5\% of reference events, so its grouping performance reflects a small matched subset. Similarly, unified Qwen3.5-0.8B attains creation recall of 50.0 with only 0.2\% coverage, making that score uninformative about discovery across the full stream. Coverage, partition agreement, and creation--reuse decisions must therefore be interpreted jointly.

\textbf{Implications.}
These results motivate two questions: whether models use the accumulated library to guide their decisions, and whether they correctly judge transformation equivalence across observations. The prevailing biases suggest complementary needs: recognizing novelty despite the availability of familiar schemas, and recognizing recurrence despite changes in objects, scenes, or event descriptions. Section~\ref{sec:training} investigates how supervision under reference, counterfactual, and student-generated library histories affects these behaviors.

\section{Learning to Maintain a Skill Library}
\label{sec:training}

We investigate three supervised fine-tuning strategies that differ in the library states presented during training. Unified models learn to jointly ground events and update the library, whereas factorized models retain a frozen event proposer and train only the text-only meta-policy. 

\subsection{Supervision Strategies}
\label{sec:supervision-strategies}
The strategies share the supervised objective
\begin{equation}
\mathcal{L}_{\mathrm{SFT}}(\theta)
=
-\mathbb{E}_{(o_t,z_{t-1},y_t^*)\sim\mathcal{D}}
\log p_{\theta}(y_t^*\mid o_t,z_{t-1}),
\end{equation}
where $o_t$ is a video chunk for unified models or a proposed event description for factorized models, and $z_{t-1}$ contains the library and unfinished-event state. The target $y_t^*$ specifies grounded events and library updates in the unified paradigm, or a library-management transaction in the factorized paradigm. Each strategy constructs a different training distribution $\mathcal{D}$.

\textbf{Oracle-history SFT} uses states obtained by replaying annotated episodes in order. Each observation is paired with the library and unfinished-event state produced by preceding reference outputs, and the target specifies the appropriate response under that history. The library persists across videos; unified training additionally uses periodic resets to provide repeated supervision for initialization. This strategy teaches correct decisions under reference histories.

\textbf{Counterfactual Library-state Rebalancing} (\textsc{CLaRe}) augments these histories with edited states while holding the observation fixed. Removing a required schema changes reuse to creation, whereas adding semantically distinct lexical distractors or removing unrelated entries should preserve the correct assignment. The factorized variant also includes equivalent-schema insertion, which changes creation to reuse, and interventions on unfinished-event state. Targets are adjusted to the edited context. Counterfactual examples branch from the reference trajectory without modifying its subsequent states, supervising both sensitivity to relevant changes and consistency under irrelevant ones.

\textbf{On-policy correction} addresses the mismatch between reference histories and states encountered during inference. We roll out the student over ordered training episodes and obtain teacher-labeled targets for the states reached through its own predictions. These targets are recomputed for the encountered library: if the student failed to introduce a required schema earlier, a recurring transformation may now require creation even when the reference history prescribes reuse. Fine-tuning on these examples provides supervision under accumulated student errors. The objective remains supervised learning; student rollouts change the training-state distribution, similar to on-policy teacher--student distillation for language models~\citep{agarwal2024policy,gu2024minillm}.

\subsection{Effects of Supervision}
\label{sec:supervision-results}

\begin{table*}[t]
\caption{\textbf{Effect of supervision on \texttt{Video2Skill} across two backbones.}
\protect\smash{\protect\colorbox{sftpink}{Pink}} highlights column maxima
among the three supervised variants within each backbone and paradigm,
including ties.
All scores are multiplied by 100; ARI can be negative.}
\label{tab:sft}
\vspace{3pt}
\centering
\small
\setlength{\tabcolsep}{4pt}
\renewcommand{\arraystretch}{1.0}
\resizebox{\textwidth}{!}{%
\begin{tabular}{lcccccccccccc}
\toprule
\multirow{2}{*}{Method}
& \multicolumn{6}{c}{\textsc{RoboInter}}
& \multicolumn{6}{c}{\textsc{HD-EPIC}} \\
\cmidrule(lr){2-7}\cmidrule(lr){8-13}
& Cov. & Pair P & Pair R & ARI & Create R & Reuse R
& Cov. & Pair P & Pair R & ARI & Create R & Reuse R \\
\midrule
\rowcolor{gray!10}
\multicolumn{13}{l}{\textit{\textbf{Unified}: a single VLM watches the video and manages the library in one pass}} \\

Qwen3.5-4B
& 56.9 & 36.7 & 63.1 & 19.4 & 20.0 & 99.3
& 61.4 & 12.5 & 32.7 & 8.7 & 11.1 & 98.5 \\
\quad + Oracle-history SFT
& 52.4 & \sftbest{67.6} & \sftbest{74.5} & \sftbest{60.0}
& \sftbest{33.3} & 97.7
& 49.6 & 31.1 & 42.8 & \sftbest{31.0} & 28.6 & 97.3 \\
\quad + \textsc{CLaRe}
& \sftbest{59.2} & 66.1 & 70.1 & 56.2 & 20.0 & 97.0
& 53.3 & \sftbest{31.8} & 37.7 & 29.2 & \sftbest{40.0} & 97.5 \\
\quad + On-policy correction
& 50.2 & 65.1 & 73.0 & 56.3 & \sftbest{33.3} & \sftbest{98.4}
& \sftbest{55.8} & 27.3 & \sftbest{46.8} & 28.2 & 19.1 & \sftbest{97.6} \\
\midrule

LLaVA-OneVision-2-8B
& 37.8 & 22.5 & 100.0 & 0.0 & 7.1 & 100.0
& 21.6 & 13.4 & 52.4 & 13.2 & 12.8 & 95.7 \\
\quad + Oracle-history SFT
& 54.1 & \sftbest{66.9} & 77.2 & \sftbest{60.9}
& \sftbest{38.5} & 98.6
& 51.6 & \sftbest{31.4} & 43.4 & \sftbest{30.8} & 34.2 & 97.2 \\
\quad + \textsc{CLaRe}
& \sftbest{60.9} & 62.4 & \sftbest{78.3} & 58.1 & 33.3 & \sftbest{98.7}
& \sftbest{55.3} & 26.4 & 42.3 & 25.9 & \sftbest{35.0} & 97.5 \\
\quad + On-policy correction
& 50.4 & 65.1 & 72.8 & 55.5 & 30.8 & 97.7
& 46.7 & 27.5 & \sftbest{45.4} & 28.0 & 30.0 & \sftbest{97.8} \\
\midrule

\rowcolor{gray!10}
\multicolumn{13}{l}{\textit{\textbf{Factorized}: the VLM first proposes events then updates the library via a text-only meta-policy}} \\

Qwen3.5-4B
& 46.6 & 31.3 & 8.9 & 3.2 & 26.7 & 68.8
& 50.1 & 6.3 & 15.9 & -1.0 & 44.2 & 47.6 \\
\quad + Oracle-history SFT
& \sftbest{46.6} & 33.8 & 27.4 & 10.4 & \sftbest{20.0} & 93.2
& \sftbest{50.1} & \sftbest{26.7} & 22.9 & 19.5 & \sftbest{39.5} & 90.5 \\
\quad + \textsc{CLaRe}
& \sftbest{46.6} & 30.5 & 15.9 & 4.7 & \sftbest{20.0} & 79.9
& \sftbest{50.1} & 25.7 & \sftbest{25.0} & \sftbest{19.9} & 34.9 & 90.5 \\
\quad + On-policy correction
& \sftbest{46.6} & \sftbest{33.9} & \sftbest{31.4}
& \sftbest{11.4} & \sftbest{20.0} & \sftbest{94.0}
& \sftbest{50.1} & 26.0 & 23.8 & 19.5 & 37.2 & \sftbest{93.5} \\
\midrule

LLaVA-OneVision-2-8B
& 71.7 & 27.4 & 0.8 & 0.1 & 80.0 & 29.1
& 40.5 & 17.2 & 0.6 & 0.7 & 87.8 & 16.0 \\
\quad + Oracle-history SFT
& \sftbest{71.7} & 30.9 & 10.1 & 2.7 & \sftbest{46.7} & 74.7
& \sftbest{40.5} & \sftbest{18.4} & 5.5 & 5.7 & \sftbest{61.0} & 60.1 \\
\quad + \textsc{CLaRe}
& \sftbest{71.7} & \sftbest{32.8} & 25.0 & \sftbest{7.5}
& 13.3 & \sftbest{94.6}
& \sftbest{40.5} & 17.8 & 10.1 & 8.7 & 46.3 & 74.0 \\
\quad + On-policy correction
& \sftbest{71.7} & 30.8 & \sftbest{25.1} & 5.5 & 20.0 & \sftbest{94.6}
& \sftbest{40.5} & 16.8 & \sftbest{13.9} & \sftbest{10.1}
& 46.3 & \sftbest{84.2} \\
\bottomrule
\end{tabular}%
}
\vspace{-20pt}
\end{table*}

We compare the three supervision strategies on Qwen3.5-4B and LLaVA-OneVision-2-8B under both paradigms and present the results in Table~\ref{tab:sft}.

\textbf{Supervision mitigates both failure tendencies.}
All three strategies improve ARI over their zero-shot counterparts across both backbones, domains, and paradigms. Unified models improve both pair precision and creation recall. For example, oracle-history SFT raises Qwen3.5-4B's pair precision from 30.0 to 67.6 on \textsc{RoboInter} and from 12.7 to 31.1 on \textsc{HD-EPIC}, indicating less mixing of distinct transformations among matched events. These gains should be interpreted alongside changes in coverage: in the latter setting, coverage falls from 69.9 to 49.6, so the partition scores concern different matched subsets. Factorized models improve recurrence grouping while coverage remains fixed. For LLaVA-OneVision-2-8B on \textsc{RoboInter}, on-policy correction raises pair recall from 0.8 to 25.1 and reuse recall from 29.1 to 94.6, while creation recall falls from 80.0 to 20.0. Training therefore alleviates fragmentation, but also causes more first occurrences of reference transformations to receive previously used schema names.

\textbf{Training-state interventions have different benefits across settings.}
Oracle-history SFT achieves the highest unified ARI in all four backbone--domain settings. Counterfactual and student-generated histories can improve other aspects of performance without improving overall partition agreement. On \textsc{HD-EPIC}, unified \textsc{CLaRe} raises Qwen3.5-4B's coverage from 49.6 to 53.3 and creation recall from 28.6 to 40.0 relative to oracle-history SFT, while ARI decreases from 31.0 to 29.2. The effects also depend on the backbone: in the factorized paradigm on \textsc{RoboInter}, \textsc{CLaRe} improves LLaVA's ARI from 2.7 to 7.5 but reduces Qwen's from 10.4 to 4.7. On-policy correction attains the highest factorized reuse recall in every setting, including one tie, yet does not consistently achieve the highest ARI. The additional training-state interventions thus provide setting-specific benefits rather than uniform improvements over reference-history supervision.

\textbf{Better grouping does not ensure correct creation and reuse.}
Despite improvements over zero-shot performance, trained unified models retain a substantial imbalance: reuse recall remains at 97.0--98.7, while creation recall reaches only 19.1--40.0. Most first occurrences of reference transformations in the matched sequence therefore still receive previously used names. Factorized models illustrate a complementary limitation: their improved reuse recall coexists with pair recall of only 5.5--31.4. Reuse recall records whether a schema name has appeared before, not whether it is the appropriate schema for the current transformation. A model can therefore reuse names frequently while selecting incorrect schemas or distributing recurring instances across duplicate entries. Supervision improves partition agreement, but reliable library maintenance still requires both recognizing novelty and preserving the identity of recurring transformations.

\section{Further Discussions}
\begin{wrapfigure}[17]{r}{0.53\textwidth}
    \centering
    \vspace{-10pt}
    \includegraphics[width=\linewidth]{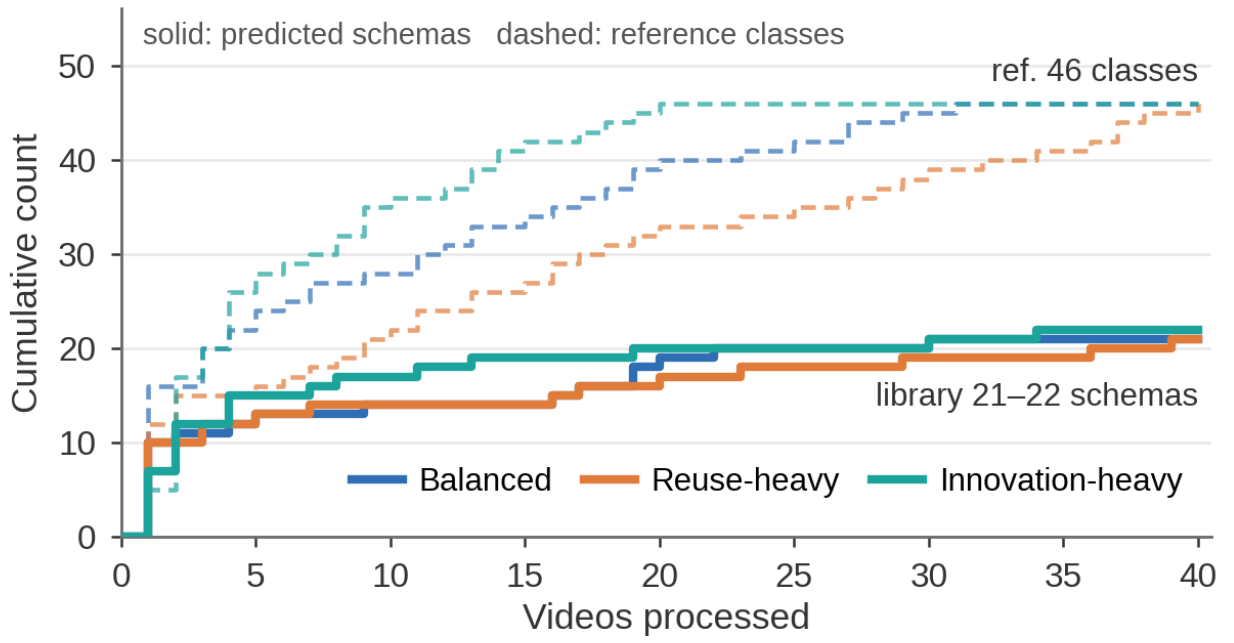}
    \vspace{-22pt}
    \caption{\textbf{Library growth under different video orders.}
    Unified Qwen3.5-4B with oracle-history SFT on the 40
    \textsc{HD-EPIC} clips.
    Solid and dashed curves show cumulative predicted schemas
    and observed reference classes, respectively.
    All three orders end with 21--22 schemas versus 46 reference classes.}
    \label{fig:order-growth}
\end{wrapfigure}
\textbf{How does video order affect skill discovery?}
We investigate the effect of presentation order by replaying the same 40 \textsc{HD-EPIC} clips as closed-loop episodes. Alongside the default balanced order, we introduce a reuse-heavy order that spreads first occurrences of reference transformations throughout the episode and an innovation-heavy order that concentrates them near the beginning. Figure~\ref{fig:order-growth} shows the growth trajectories for oracle-history SFT. Despite different trajectories, the libraries end with similar sizes of 21--22 schemas, compared with 46 reference classes. Under the innovation-heavy order, all 46 reference classes have appeared by the twentieth video, yet the model has introduced only 20 schemas. Under the other orders, reference classes continue to accumulate later while the predicted libraries expand slowly. Making novelty appear earlier therefore does not substantially enlarge the model's final repertoire. Beyond library size, oracle-history SFT achieves a mean pairwise Jaccard similarity of 0.78 between final schema-name sets across orders, compared with 0.59 for the zero-shot model; \textsc{CLaRe} also achieves a mean of 0.78. Training thus improves consistency in the names retained across orders, while limited library expansion persists. 

\textbf{Are training gains shared across skills?}
We group training-seen reference classes by their frequency in the training annotations and report class-averaged pair F1 (Table~\ref{tab:class-analysis}a). High-frequency classes benefit consistently from supervision: their scores rise from 19.0 to 34.9--40.0 on \textsc{HD-EPIC} and from 25.5 to 40.9--51.7 on \textsc{RoboInter}. Improvements elsewhere are smaller and depend on the training strategy. Medium-frequency classes remain below 10 on both datasets, while low-frequency classes improve under oracle-history SFT and on-policy correction but not consistently under \textsc{CLaRe}. Performance also does not vary monotonically with frequency: low-frequency groups sometimes outperform medium-frequency groups, including under oracle-history SFT and on-policy correction on both datasets. The results therefore support uneven benefits across the skill repertoire, with the most consistent gains among frequently demonstrated transformations, but do not establish a simple relationship between training frequency and grouping quality.

\textbf{Do models recognize unseen transformations as novel?}
Eleven \textsc{HD-EPIC} test classes, comprising 59 segments, are absent from task-specific training annotations. Their temporal coverage after supervision is 57.6--66.1, exceeding the corresponding seen-class coverage under every strategy, as presented in Table~\ref{tab:class-analysis}b. However, this recovery rarely leads to creation: trained models assign a previously unused schema name at the first matched occurrence of only zero or one of the nine to ten covered unseen classes. Seen-class creation recall, in contrast, improves from 14.7 to 34.4--48.4. Higher grouping scores do not resolve this gap: oracle-history SFT raises unseen-class macro pair F1 from 8.6 to 25.5, yet creates a new name at the first matched occurrence of only one of ten covered classes. Improved grouping therefore does not establish novelty recognition, nor does failure to create at the first matched occurrence establish that a schema is never introduced later. These results highlight the difficulty of recognizing when existing schemas are insufficient, even among events that are successfully matched in time.\looseness=-1

\begin{table}[t]
\centering
\caption{\textbf{Skill grouping and novelty recognition with unified Qwen3.5-4B.}
(a) Macro pair F1 by training frequency (class counts in parentheses).
(b) Classes seen or unseen in task-specific training.
For unseen classes, $k/n$ counts those receiving a new schema name
at their first matched occurrence, out of all covered classes.
Other scores are multiplied by 100.}
\vspace{-5pt}
\label{tab:class-analysis}

\begingroup
\fontsize{9}{11}\selectfont
\setlength{\tabcolsep}{2.5pt}
\renewcommand{\arraystretch}{0.9}
\setlength{\heavyrulewidth}{0.7pt}
\setlength{\lightrulewidth}{0.35pt}
\setlength{\aboverulesep}{1.2pt}
\setlength{\belowrulesep}{1.2pt}

\newcommand{\classLeftStrut}{\rule[-3pt]{0pt}{10.5pt}}
\newcommand{\classRightStrut}{\rule[-3.6pt]{0pt}{12.6pt}}

\begin{minipage}[t]{0.475\linewidth}
\vspace{0pt}
\centering
\begin{tabular*}{\linewidth}{@{\extracolsep{\fill}}lcccc@{}}
\toprule
\makecell[l]{Frequency}
& \makecell[c]{Zero-\\shot}
& \makecell[c]{Oracle\\history}
& \makecell[c]{\textsc{CLaRe}}
& \makecell[c]{On-policy\\correction} \\
\midrule
\rowcolor{gray!8}[0pt][0pt]
\multicolumn{5}{@{}l@{}}{\textsc{HD-EPIC}} \\
\classLeftStrut High (11) & 19.0 & 38.3 & 34.9 & 40.0 \\
\classLeftStrut Mid (13)  & 4.2  & 7.5  & 9.7  & 7.0  \\
\classLeftStrut Low (11)  & 8.5  & 15.4 & 7.9  & 17.3 \\
\midrule
\rowcolor{gray!8}[0pt][0pt]
\multicolumn{5}{@{}l@{}}{\textsc{RoboInter}} \\
\classLeftStrut High (5) & 25.5 & 45.0 & 40.9 & 51.7 \\
\classLeftStrut Mid (5)  & 2.5  & 1.5  & 1.7  & 2.9  \\
\classLeftStrut Low (5)  & 3.8  & 5.8  & 5.1  & 7.1  \\
\bottomrule
\end{tabular*}
\par\vspace{3pt}
\textbf{(a) Grouping by training frequency}
\end{minipage}%
\hspace{0.02\linewidth}%
\begin{minipage}[t]{0.50\linewidth}
\vspace{0pt}
\centering
\begin{tabular*}{\linewidth}{@{\extracolsep{\fill}}lcccc@{}}
\toprule
\makecell[l]{Metric}
& \makecell[c]{Zero-\\shot}
& \makecell[c]{Oracle\\history}
& \makecell[c]{\textsc{CLaRe}}
& \makecell[c]{On-policy\\correction} \\
\midrule
\rowcolor{gray!8}[0pt][0pt]
\multicolumn{5}{@{}l@{}}{Training-seen} \\
\classRightStrut Coverage & 61.3 & 48.7 & 53.1 & 51.8 \\
\classRightStrut Create R & 14.7 & 34.4 & 48.4 & 34.4 \\
\midrule
\rowcolor{gray!8}[0pt][0pt]
\multicolumn{5}{@{}l@{}}{Training-unseen} \\
\classRightStrut Coverage
& 62.7 & 66.1 & 57.6 & 66.1 \\
\classRightStrut Macro pair F1
& 8.6 & 25.5 & 23.2 & 10.2 \\
\classRightStrut Create ($k/n$)
& $0/11$ & $1/10$ & $1/9$ & $0/10$ \\
\bottomrule
\end{tabular*}
\par\vspace{3pt}
\textbf{(b) Seen vs.\ unseen on \textsc{HD-EPIC}}
\end{minipage}
\endgroup
\vspace{-18pt}
\end{table}

\section{Related Work}
\vspace{-2pt}
\textbf{Video understanding and streaming models.}
Action recognition, temporal localization, and procedural understanding typically evaluate predictions against predefined action or keystep categories~\citep{damen2022rescaling,grauman2022ego4d,grauman2024ego,perrett2025hd,sener2022assembly101,zhukov2019cross,zhang2026progresslm,zhang2026progress,zhang2026lens,liu2026scar}, while video QA and long-video benchmarks assess responses to supplied questions~\citep{xiao2021next,mangalam2023egoschema,fu2025video,wu2024longvideobench,zhang2025vlm2}. Unsupervised action segmentation goes beyond predefined categories by discovering recurring action structure, typically through offline clustering over complete sequences~\citep{sener2018unsupervised,kukleva2019unsupervised}. Online action detection and streaming video models support incremental processing and temporal memory~\citep{xu2019temporal,xu2021long,zhang2025flash,song2024moviechat,he2024ma,zhang2026streamscout}, with dedicated benchmarks evaluating understanding as observations arrive~\citep{lin2026streamingbench,niu2025ovo}. 
\texttt{Video2Skill} focuses on how recurring events are consolidated into reusable transformation schemas. Models assign temporally grounded events to a library, deciding when to reuse an existing schema or introduce a new one, with the resulting library conditioning subsequent predictions.\looseness=-1

\textbf{Skill discovery and evolving concept inventories.}
Reinforcement learning studies temporally extended behaviors acquired through interaction~\citep{sutton1999between,bacon2017option,eysenbach2019diversity,sharma2020dynamics,park2024metra}, while imitation and continual learning infer skills or expand behavioral repertoires from demonstrations~\citep{pertsch2021accelerating,ajay2021opal,zhu2022bottom,wan2024lotus,zhang2023bootstrap,zhang2026evolving}. XSkill~\citep{xu2023xskill} and UniSkill~\citep{kim2025uniskill} learn shared skill representations from human and robot videos for downstream control; XSkill uses a fixed-size set of learned prototypes. LLM agents can grow explicit skill libraries online, typically using execution and environment feedback to validate new skills~\citep{wang2024voyager,wang2025jarvis,wang2025agent,zheng2025skillweaver,luo2026pyspatial}, and research agents can evolve reusable procedures through proactive online exploration~\citep{wang2026explore}. Related approaches develop symbolic models for composing robot skills, as in SkillWrapper~\citep{yang2025skillwrapper}, or build video-derived repositories that support planning and skill expansion, as in Uni-Skill~\citep{xie2026uniskill}. 
The decision to expand a library also connects to novel and generalized category discovery~\citep{han2019learning,vaze2022generalized} and their continual variants~\citep{zhang2022grow,wu2023metagcd,ma2024happy}, as well as to continual learning~\citep{zhang2024core,rong2025can} and in-context learning~\citep{brown2020language,pi2025personalized}: creating a schema accommodates a new concept, while reusing one exploits concepts already held in context. \texttt{Video2Skill} studies this familiar-versus-novel distinction for temporally grounded transformations with explicit argument slots, allowing instances involving different objects and scenes to share a schema.\looseness=-1

\section{Conclusion}
\label{sec:conclusion}

We introduced \texttt{Video2Skill} to study whether models can turn streaming visual experience into a persistent library of reusable skills. Across robot manipulation and egocentric kitchen activity, evaluations of 19 VLMs reveal complementary weaknesses: unified models tend to merge distinct transformations, while factorized models frequently assign recurring transformations to duplicate schemas. Oracle-history SFT, counterfactual library-state rebalancing, and on-policy correction partially alleviate these failures, but do not reliably resolve the boundary between novelty and reuse. Our findings underscore the need to evaluate event coverage and abstraction quality jointly, and establish a testbed for learning coherent skill libraries that remain useful as experience accumulates.

\textbf{Limitations.} \texttt{Video2Skill} focuses on robot manipulation and egocentric kitchen activity. Although these domains provide varied objects and transformations, broader settings such as outdoor activitie remain unexplored. Our skill schemas describe observed physical transformations; their utility for downstream planning and executable control requires further evaluation. \looseness=-1

% \section*{AI Use Statement}
% We used generative AI tools to assist with manuscript drafting, translation, language refinement, literature search, and figure and table preparation. AI assistance also included selected code implementation, interpretation of experimental results, and feedback on the design of additional analyses. The research was led by the authors, who determined the research questions and made all final methodological and experimental decisions. The authors reviewed the AI-assisted material and take full responsibility for the final manuscript, including its claims, experimental results, and accompanying code. \looseness=-1

\bibliography{iclr2027_conference}

@inproceedings{chen2024internvl,
  title     = {{InternVL}: Scaling up Vision Foundation Models and Aligning for Generic Visual-Linguistic Tasks},
  author    = {Chen, Zhe and Wu, Jiannan and Wang, Wenhai and Su, Weijie and Chen, Guo and Xing, Sen and Zhong, Muyan and Zhang, Qinglong and Zhu, Xizhou and Lu, Lewei and Li, Bin and Luo, Ping and Lu, Tong and Qiao, Yu and Dai, Jifeng},
  booktitle = {Proceedings of the IEEE/CVF Conference on Computer Vision and Pattern Recognition},
  pages     = {24185--24198},
  year      = {2024}
}

@article{zhang2025video,
  title   = {Video Instruction Tuning with Synthetic Data},
  author  = {Zhang, Yuanhan and Wu, Jinming and Li, Wei and Li, Bo and Ma, Zejun and Liu, Ziwei and Li, Chunyuan},
  journal = {Transactions on Machine Learning Research},
  year    = {2025},
  url     = {https://openreview.net/forum?id=8Livf4oZxz}
}

@article{bai2025qwen2,
  title   = {{Qwen2.5-VL} Technical Report},
  author  = {Bai, Shuai and Chen, Keqin and Liu, Xuejing and Wang, Jialin and Ge, Wenbin and Song, Sibo and Dang, Kai and Wang, Peng and Wang, Shijie and Tang, Jun and Zhong, Humen and Zhu, Yuanzhi and Yang, Mingkun and Li, Zhaohai and Wan, Jianqiang and Wang, Pengfei and Ding, Wei and Fu, Zheren and Xu, Yiheng and Ye, Jiabo and Zhang, Xi and Xie, Tianbao and Cheng, Zesen and Zhang, Hang and Yang, Zhibo and Xu, Haiyang and Lin, Junyang},
  journal = {arXiv preprint arXiv:2502.13923},
  year    = {2025}
}

@article{yu2026memora,
  title   = {{MEMORA}: Embodied Action Memory from Egocentric Videos for Reasoning and Planning},
  author  = {Yu, Zihao and Yuan, Xiu and Zhang, Chongjie},
  journal = {arXiv preprint arXiv:2607.14252},
  year    = {2026}
}

@article{yang2025skillwrapper,
  title   = {{SkillWrapper}: Generative Predicate Invention for Task-level Robot Planning},
  author  = {Yang, Ziyi and Hedegaard, Benned and Jaafar, Ahmed and Wei, Yichen and Thompson, Skye and Raman, Shreyas S. and Fu, Haotian and Tellex, Stefanie and Konidaris, George and Paulius, David and Shah, Naman},
  journal = {arXiv preprint arXiv:2511.18203},
  year    = {2025}
}

@article{liu2026scar,
  title={SCAR: Self-Supervised Continuous Action Representation Learning},
  author={Liu, Hongjia and Feng, Fan and Fu, Minghao and Wang, Xinyue and Lu, Haofei and Huang, Biwei},
  journal={arXiv preprint arXiv:2605.16412},
  year={2026}
}

@inproceedings{xie2026uniskill,
  title     = {{Uni-Skill}: Building Self-Evolving Skill Repository for Generalizable Robotic Manipulation},
  author    = {Xie, Senwei and Zhang, Yuntian and Wang, Ruiping and Chen, Xilin},
  booktitle = {IEEE International Conference on Robotics and Automation},
  year      = {2026}
}

@article{wang2024voyager,
  title   = {{Voyager}: An Open-Ended Embodied Agent with Large Language Models},
  author  = {Wang, Guanzhi and Xie, Yuqi and Jiang, Yunfan and Mandlekar, Ajay and Xiao, Chaowei and Zhu, Yuke and Fan, Linxi and Anandkumar, Anima},
  journal = {Transactions on Machine Learning Research},
  year    = {2024},
  url     = {https://openreview.net/forum?id=ehfRiF0R3a}
}

@inproceedings{ichter2023can,
  title     = {Do As {I} Can, Not As {I} Say: Grounding Language in Robotic Affordances},
  author    = {Ichter, Brian and Brohan, Anthony and Chebotar, Yevgen and Finn, Chelsea and Hausman, Karol and Herzog, Alexander and Ho, Daniel and Ibarz, Julian and Irpan, Alex and Jang, Eric and Julian, Ryan and Kalashnikov, Dmitry and Levine, Sergey and Lu, Yao and Parada, Carolina and Rao, Kanishka and Sermanet, Pierre and Toshev, Alexander T. and Vanhoucke, Vincent and Xia, Fei and Xiao, Ted and Xu, Peng and Yan, Mengyuan and Brown, Noah and Ahn, Michael and Cortes, Omar and Sievers, Nicolas and Tan, Clayton and Xu, Sichun and Reyes, Diego and Rettinghouse, Jarek and Quiambao, Jornell and Pastor, Peter and Luu, Linda and Lee, Kuang-Huei and Kuang, Yuheng and Jesmonth, Sally and Joshi, Nikhil J. and Jeffrey, Kyle and Ruano, Rosario Jauregui and Hsu, Jasmine and Gopalakrishnan, Keerthana and David, Byron and Zeng, Andy and Fu, Chuyuan Kelly},
  booktitle = {Proceedings of The 6th Conference on Robot Learning},
  series    = {Proceedings of Machine Learning Research},
  volume    = {205},
  pages     = {287--318},
  publisher = {PMLR},
  year      = {2023},
  url       = {https://proceedings.mlr.press/v205/ichter23a.html}
}

@inproceedings{liang2023code,
  title     = {Code as Policies: Language Model Programs for Embodied Control},
  author    = {Liang, Jacky and Huang, Wenlong and Xia, Fei and Xu, Peng and Hausman, Karol and Ichter, Brian and Florence, Pete and Zeng, Andy},
  booktitle = {IEEE International Conference on Robotics and Automation},
  pages     = {9493--9500},
  publisher = {IEEE},
  year      = {2023}
}

@inproceedings{wan2024lotus,
  title     = {{LOTUS}: Continual Imitation Learning for Robot Manipulation Through Unsupervised Skill Discovery},
  author    = {Wan, Weikang and Zhu, Yifeng and Shah, Rutav and Zhu, Yuke},
  booktitle = {IEEE International Conference on Robotics and Automation},
  pages     = {537--544},
  publisher = {IEEE},
  year      = {2024}
}

@inproceedings{zhang2023bootstrap,
  title     = {Bootstrap Your Own Skills: Learning to Solve New Tasks with Large Language Model Guidance},
  author    = {Zhang, Jesse and Zhang, Jiahui and Pertsch, Karl and Liu, Ziyi and Ren, Xiang and Chang, Minsuk and Sun, Shao-Hua and Lim, Joseph J.},
  booktitle = {Proceedings of the 7th Conference on Robot Learning},
  series    = {Proceedings of Machine Learning Research},
  volume    = {229},
  pages     = {302--325},
  publisher = {PMLR},
  year      = {2023}
}

@inproceedings{perrett2025hd,
  title     = {{HD-EPIC}: A Highly-Detailed Egocentric Video Dataset},
  author    = {Perrett, Toby and Darkhalil, Ahmad and Sinha, Saptarshi and Emara, Omar and Pollard, Sam and Parida, Kranti Kumar and Liu, Kaiting and Gatti, Prajwal and Bansal, Siddhant and Flanagan, Kevin and Chalk, Jacob and Zhu, Zhifan and Guerrier, Rhodri and Abdelazim, Fahd and Zhu, Bin and Moltisanti, Davide and Wray, Michael and Doughty, Hazel and Damen, Dima},
  booktitle = {Proceedings of the IEEE/CVF Conference on Computer Vision and Pattern Recognition},
  pages     = {23901--23913},
  year      = {2025}
}

@inproceedings{khazatsky2024droid,
  title     = {{DROID}: A Large-Scale In-The-Wild Robot Manipulation Dataset},
  author    = {Khazatsky, Alexander and Pertsch, Karl and Nair, Suraj and Balakrishna, Ashwin and Dasari, Sudeep and Karamcheti, Siddharth and Nasiriany, Soroush and Srirama, Mohan Kumar and Chen, Lawrence Yunliang and Ellis, Kirsty and Fagan, Peter David and Hejna, Joey and Itkina, Masha and Lepert, Marion and Ma, Yecheng Jason and Miller, Patrick Tree and Wu, Jimmy and Belkhale, Suneel and Dass, Shivin and Ha, Huy and Jain, Arhan and Lee, Abraham and Lee, Youngwoon and Memmel, Marius and Park, Sungjae and Radosavovic, Ilija and Wang, Kaiyuan and Zhan, Albert and Black, Kevin and Chi, Cheng and Hatch, Kyle Beltran and Lin, Shan and Lu, Jingpei and Mercat, Jean and Rehman, Abdul and Sanketi, Pannag R. and Sharma, Archit and Simpson, Cody and Vuong, Quan and Walke, Homer Rich and Wulfe, Blake and Xiao, Ted and Yang, Jonathan Heewon and Yavary, Arefeh and Zhao, Tony Z. and Agia, Christopher and Baijal, Rohan and Castro, Mateo Guaman and Chen, Daphne and Chen, Qiuyu and Chung, Trinity and Drake, Jaimyn and Foster, Ethan Paul and Gao, Jensen and Herrera, David Antonio and Heo, Minho and Hsu, Kyle and Hu, Jiaheng and Jackson, Donovon and Le, Charlotte and Li, Yunshuang and Lin, Roy and Ma, Zehan and Maddukuri, Abhiram and Mirchandani, Suvir and Morton, Daniel and Nguyen, Tony and O'Neill, Abigail and Scalise, Rosario and Seale, Derick and Son, Victor and Tian, Stephen and Tran, Emi and Wang, Andrew E. and Wu, Yilin and Xie, Annie and Yang, Jingyun and Yin, Patrick and Zhang, Yunchu and Bastani, Osbert and Berseth, Glen and Bohg, Jeannette and Goldberg, Ken and Gupta, Abhinav and Gupta, Abhishek and Jayaraman, Dinesh and Lim, Joseph J. and Malik, Jitendra and Mart{\'i}n-Mart{\'i}n, Roberto and Ramamoorthy, Subramanian and Sadigh, Dorsa and Song, Shuran and Wu, Jiajun and Yip, Michael C. and Zhu, Yuke and Kollar, Thomas and Levine, Sergey and Finn, Chelsea},
  booktitle = {Proceedings of Robotics: Science and Systems},
  address   = {Delft, Netherlands},
  month     = jul,
  year      = {2024}
}

@inproceedings{fang2024rh20t,
  title     = {{RH20T}: A Comprehensive Robotic Dataset for Learning Diverse Skills in One-Shot},
  author    = {Fang, Hao-Shu and Fang, Hongjie and Tang, Zhenyu and Liu, Jirong and Wang, Chenxi and Wang, Junbo and Zhu, Haoyi and Lu, Cewu},
  booktitle = {2024 IEEE International Conference on Robotics and Automation (ICRA)},
  pages     = {653--660},
  year      = {2024}
}

@article{damen2022rescaling,
  title   = {Rescaling Egocentric Vision: Collection, Pipeline and Challenges for {EPIC-KITCHENS-100}},
  author  = {Damen, Dima and Doughty, Hazel and Farinella, Giovanni Maria and Furnari, Antonino and Ma, Jian and Kazakos, Evangelos and Moltisanti, Davide and Munro, Jonathan and Perrett, Toby and Price, Will and Wray, Michael},
  journal = {International Journal of Computer Vision},
  volume  = {130},
  pages   = {33--55},
  year    = {2022}
}

@inproceedings{grauman2022ego4d,
  title     = {{Ego4D}: Around the World in 3,000 Hours of Egocentric Video},
  author    = {Grauman, Kristen and Westbury, Andrew and Byrne, Eugene and Chavis, Zachary and Furnari, Antonino and Girdhar, Rohit and Hamburger, Jackson and Jiang, Hao and Liu, Miao and Liu, Xingyu and Martin, Miguel and Nagarajan, Tushar and Radosavovic, Ilija and Ramakrishnan, Santhosh Kumar and Ryan, Fiona and Sharma, Jayant and Wray, Michael and Xu, Mengmeng and Xu, Eric Zhongcong and Zhao, Chen and Bansal, Siddhant and Batra, Dhruv and Cartillier, Vincent and Crane, Sean and Do, Tien and Doulaty, Morrie and Erapalli, Akshay and Feichtenhofer, Christoph and Fragomeni, Adriano and Fu, Qichen and Gebreselasie, Abrham and Gonz{\'a}lez, Cristina and Hillis, James and Huang, Xuhua and Huang, Yifei and Jia, Wenqi and Khoo, Weslie and Kol{\'a}{\v r}, J{\'a}chym and Kottur, Satwik and Kumar, Anurag and Landini, Federico and Li, Chao and Li, Yanghao and Li, Zhenqiang and Mangalam, Karttikeya and Modhugu, Raghava and Munro, Jonathan and Murrell, Tullie and Nishiyasu, Takumi and Price, Will and Ruiz, Paola and Ramazanova, Merey and Sari, Leda and Somasundaram, Kiran and Southerland, Audrey and Sugano, Yusuke and Tao, Ruijie and Vo, Minh and Wang, Yuchen and Wu, Xindi and Yagi, Takuma and Zhao, Ziwei and Zhu, Yunyi and Arbel{\'a}ez, Pablo and Crandall, David and Damen, Dima and Farinella, Giovanni Maria and Fuegen, Christian and Ghanem, Bernard and Ithapu, Vamsi Krishna and Jawahar, C. V. and Joo, Hanbyul and Kitani, Kris and Li, Haizhou and Newcombe, Richard and Oliva, Aude and Park, Hyun Soo and Rehg, James M. and Sato, Yoichi and Shi, Jianbo and Shou, Mike Zheng and Torralba, Antonio and Torresani, Lorenzo and Yan, Mingfei and Malik, Jitendra},
  booktitle = {Proceedings of the IEEE/CVF Conference on Computer Vision and Pattern Recognition},
  pages     = {18995--19012},
  year      = {2022}
}

@inproceedings{sener2022assembly101,
  title     = {{Assembly101}: A Large-Scale Multi-View Video Dataset for Understanding Procedural Activities},
  author    = {Sener, Fadime and Chatterjee, Dibyadip and Shelepov, Daniel and He, Kun and Singhania, Dipika and Wang, Robert and Yao, Angela},
  booktitle = {Proceedings of the IEEE/CVF Conference on Computer Vision and Pattern Recognition},
  pages     = {21096--21106},
  year      = {2022}
}

@inproceedings{zhukov2019cross,
  title     = {Cross-Task Weakly Supervised Learning from Instructional Videos},
  author    = {Zhukov, Dimitri and Alayrac, Jean-Baptiste and Cinbis, Ramazan Gokberk and Fouhey, David and Laptev, Ivan and Sivic, Josef},
  booktitle = {Proceedings of the IEEE/CVF Conference on Computer Vision and Pattern Recognition},
  pages     = {3537--3545},
  year      = {2019}
}

@inproceedings{grauman2024ego,
  title     = {{Ego-Exo4D}: Understanding Skilled Human Activity from First- and Third-Person Perspectives},
  author    = {Grauman, Kristen and Westbury, Andrew and Torresani, Lorenzo and Kitani, Kris and Malik, Jitendra and Afouras, Triantafyllos and Ashutosh, Kumar and Baiyya, Vijay and Bansal, Siddhant and Boote, Bikram and others},
  booktitle = {Proceedings of the IEEE/CVF Conference on Computer Vision and Pattern Recognition},
  pages     = {19383--19400},
  year      = {2024}
}

@inproceedings{xiao2021next,
  title     = {{NExT-QA}: Next Phase of Question-Answering to Explaining Temporal Actions},
  author    = {Xiao, Junbin and Shang, Xindi and Yao, Angela and Chua, Tat-Seng},
  booktitle = {Proceedings of the IEEE/CVF Conference on Computer Vision and Pattern Recognition},
  pages     = {9777--9786},
  year      = {2021}
}

@inproceedings{mangalam2023egoschema,
  title     = {{EgoSchema}: A Diagnostic Benchmark for Very Long-Form Video Language Understanding},
  author    = {Mangalam, Karttikeya and Akshulakov, Raiymbek and Malik, Jitendra},
  booktitle = {Advances in Neural Information Processing Systems},
  volume    = {36},
  pages     = {46212--46244},
  year      = {2023}
}

@inproceedings{fu2025video,
  title     = {{Video-MME}: The First-Ever Comprehensive Evaluation Benchmark of Multi-Modal {LLMs} in Video Analysis},
  author    = {Fu, Chaoyou and Dai, Yuhan and Luo, Yongdong and Li, Lei and Ren, Shuhuai and Zhang, Renrui and Wang, Zihan and Zhou, Chenyu and Shen, Yunhang and Zhang, Mengdan and others},
  booktitle = {Proceedings of the IEEE/CVF Conference on Computer Vision and Pattern Recognition},
  pages     = {24108--24118},
  year      = {2025}
}

@inproceedings{wu2024longvideobench,
  title     = {{LongVideoBench}: A Benchmark for Long-Context Interleaved Video-Language Understanding},
  author    = {Wu, Haoning and Li, Dongxu and Chen, Bei and Li, Junnan},
  booktitle = {Advances in Neural Information Processing Systems},
  volume    = {37},
  pages     = {28828--28857},
  year      = {2024}
}

@inproceedings{sener2018unsupervised,
  title     = {Unsupervised Learning and Segmentation of Complex Activities from Video},
  author    = {Sener, Fadime and Yao, Angela},
  booktitle = {Proceedings of the IEEE Conference on Computer Vision and Pattern Recognition},
  pages     = {8368--8376},
  year      = {2018}
}

@inproceedings{kukleva2019unsupervised,
  title     = {Unsupervised Learning of Action Classes with Continuous Temporal Embedding},
  author    = {Kukleva, Anna and Kuehne, Hilde and Sener, Fadime and Gall, Juergen},
  booktitle = {Proceedings of the IEEE/CVF Conference on Computer Vision and Pattern Recognition},
  pages     = {12066--12074},
  year      = {2019}
}

@inproceedings{xu2019temporal,
  title     = {Temporal Recurrent Networks for Online Action Detection},
  author    = {Xu, Mingze and Gao, Mingfei and Chen, Yi-Ting and Davis, Larry S. and Crandall, David J.},
  booktitle = {Proceedings of the IEEE/CVF International Conference on Computer Vision},
  pages     = {5532--5541},
  year      = {2019}
}

@inproceedings{xu2021long,
  title     = {Long Short-Term Transformer for Online Action Detection},
  author    = {Xu, Mingze and Xiong, Yuanjun and Chen, Hao and Li, Xinyu and Xia, Wei and Tu, Zhuowen and Soatto, Stefano},
  booktitle = {Advances in Neural Information Processing Systems},
  volume    = {34},
  pages     = {1086--1099},
  year      = {2021}
}

@inproceedings{zhang2025flash,
  title     = {{Flash-VStream}: Efficient Real-Time Understanding for Long Video Streams},
  author    = {Zhang, Haoji and Wang, Yiqin and Tang, Yansong and Liu, Yong and Feng, Jiashi and Jin, Xiaojie},
  booktitle = {Proceedings of the IEEE/CVF International Conference on Computer Vision},
  pages     = {21059--21069},
  year      = {2025}
}

@inproceedings{song2024moviechat,
  title     = {{MovieChat}: From Dense Token to Sparse Memory for Long Video Understanding},
  author    = {Song, Enxin and Chai, Wenhao and Wang, Guanhong and Zhang, Yucheng and Zhou, Haoyang and Wu, Feiyang and Chi, Haozhe and Guo, Xun and Ye, Tian and Zhang, Yanting and others},
  booktitle = {Proceedings of the IEEE/CVF Conference on Computer Vision and Pattern Recognition},
  pages     = {18221--18232},
  year      = {2024}
}

@inproceedings{he2024ma,
  title     = {{MA-LMM}: Memory-Augmented Large Multimodal Model for Long-Term Video Understanding},
  author    = {He, Bo and Li, Hengduo and Jang, Young Kyun and Jia, Menglin and Cao, Xuefei and Shah, Ashish and Shrivastava, Abhinav and Lim, Ser-Nam},
  booktitle = {Proceedings of the IEEE/CVF Conference on Computer Vision and Pattern Recognition},
  pages     = {13504--13514},
  year      = {2024}
}

@inproceedings{qian2025dispider,
  title     = {Dispider: Enabling Video {LLMs} with Active Real-Time Interaction via Disentangled Perception, Decision, and Reaction},
  author    = {Qian, Rui and Ding, Shuangrui and Dong, Xiaoyi and Zhang, Pan and Zang, Yuhang and Cao, Yuhang and Lin, Dahua and Wang, Jiaqi},
  booktitle = {Proceedings of the IEEE/CVF Conference on Computer Vision and Pattern Recognition},
  pages     = {24045--24055},
  year      = {2025}
}

@inproceedings{lin2026streamingbench,
  title     = {{StreamingBench}: Assessing the Gap for {MLLMs} to Achieve Streaming Video Understanding},
  author    = {Lin, Junming and Fang, Zheng and Chen, Chi and Cheng, Haoxuan and Wan, Zihao and Luo, Fuwen and Wang, Ziyue and Li, Peng and Liu, Yang and Sun, Maosong},
  booktitle = {IEEE International Conference on Acoustics, Speech and Signal Processing},
  pages     = {12147--12151},
  year      = {2026}
}

@inproceedings{niu2025ovo,
  title     = {{OVO-Bench}: How Far is Your {Video-LLMs} from Real-World Online Video Understanding?},
  author    = {Niu, Junbo and Li, Yifei and Miao, Ziyang and Ge, Chunjiang and Zhou, Yuanhang and He, Qihao and Dong, Xiaoyi and Duan, Haodong and Ding, Shuangrui and Qian, Rui and Zhang, Pan and Zang, Yuhang and Cao, Yuhang and He, Conghui and Wang, Jiaqi},
  booktitle = {Proceedings of the IEEE/CVF Conference on Computer Vision and Pattern Recognition},
  pages     = {18902--18913},
  year      = {2025}
}

@article{sutton1999between,
  title   = {Between {MDPs} and Semi-{MDPs}: A Framework for Temporal Abstraction in Reinforcement Learning},
  author  = {Sutton, Richard S. and Precup, Doina and Singh, Satinder},
  journal = {Artificial Intelligence},
  volume  = {112},
  number  = {1-2},
  pages   = {181--211},
  year    = {1999}
}

@inproceedings{bacon2017option,
  title     = {The Option-Critic Architecture},
  author    = {Bacon, Pierre-Luc and Harb, Jean and Precup, Doina},
  booktitle = {Proceedings of the Thirty-First AAAI Conference on Artificial Intelligence},
  pages     = {1726--1734},
  year      = {2017}
}

@inproceedings{eysenbach2019diversity,
  title     = {Diversity is All You Need: Learning Skills without a Reward Function},
  author    = {Eysenbach, Benjamin and Gupta, Abhishek and Ibarz, Julian and Levine, Sergey},
  booktitle = {International Conference on Learning Representations},
  year      = {2019}
}

@inproceedings{
sharma2020dynamics,
title={Dynamics-Aware Unsupervised Discovery of Skills},
author={Archit Sharma and Shixiang Gu and Sergey Levine and Vikash Kumar and Karol Hausman},
booktitle={International Conference on Learning Representations},
year={2020},
url={https://openreview.net/forum?id=HJgLZR4KvH}
}

@inproceedings{park2024metra,
  title     = {{METRA}: Scalable Unsupervised {RL} with Metric-Aware Abstraction},
  author    = {Park, Seohong and Rybkin, Oleh and Levine, Sergey},
  booktitle = {International Conference on Learning Representations},
  pages     = {18579--18603},
  year      = {2024}
}

@inproceedings{pertsch2021accelerating,
  title     = {Accelerating Reinforcement Learning with Learned Skill Priors},
  author    = {Pertsch, Karl and Lee, Youngwoon and Lim, Joseph J.},
  booktitle = {Proceedings of the 4th Conference on Robot Learning},
  series    = {Proceedings of Machine Learning Research},
  volume    = {155},
  pages     = {188--204},
  publisher = {PMLR},
  year      = {2021}
}

@inproceedings{
ajay2021opal,
title={{\{}OPAL{\}}: Offline Primitive Discovery for Accelerating Offline Reinforcement Learning},
author={Anurag Ajay and Aviral Kumar and Pulkit Agrawal and Sergey Levine and Ofir Nachum},
booktitle={International Conference on Learning Representations},
year={2021},
url={https://openreview.net/forum?id=V69LGwJ0lIN}
}

@article{zhu2022bottom,
  title   = {Bottom-Up Skill Discovery from Unsegmented Demonstrations for Long-Horizon Robot Manipulation},
  author  = {Zhu, Yifeng and Stone, Peter and Zhu, Yuke},
  journal = {IEEE Robotics and Automation Letters},
  volume  = {7},
  number  = {2},
  pages   = {4126--4133},
  year    = {2022}
}

@inproceedings{xu2023xskill,
  title     = {{XSkill}: Cross Embodiment Skill Discovery},
  author    = {Xu, Mengda and Xu, Zhenjia and Chi, Cheng and Veloso, Manuela and Song, Shuran},
  booktitle = {Proceedings of the 7th Conference on Robot Learning},
  series    = {Proceedings of Machine Learning Research},
  volume    = {229},
  pages     = {3536--3555},
  publisher = {PMLR},
  year      = {2023}
}

@article{wang2025jarvis,
  title   = {{JARVIS-1}: Open-World Multi-Task Agents with Memory-Augmented Multimodal Language Models},
  author  = {Wang, Zihao and Cai, Shaofei and Liu, Anji and Jin, Yonggang and Hou, Jinbing and Zhang, Bowei and Lin, Haowei and He, Zhaofeng and Zheng, Zilong and Yang, Yaodong and Ma, Xiaojian and Liang, Yitao},
  journal = {IEEE Transactions on Pattern Analysis and Machine Intelligence},
  volume  = {47},
  number  = {3},
  pages   = {1894--1907},
  year    = {2025}
}

@inproceedings{wang2025agent,
  title     = {Agent Workflow Memory},
  author    = {Wang, Zora Zhiruo and Mao, Jiayuan and Fried, Daniel and Neubig, Graham},
  booktitle = {Proceedings of the 42nd International Conference on Machine Learning},
  series    = {Proceedings of Machine Learning Research},
  volume    = {267},
  pages     = {63897--63911},
  publisher = {PMLR},
  year      = {2025}
}

@article{zheng2025skillweaver,
  title   = {{SkillWeaver}: Web Agents Can Self-Improve by Discovering and Honing Skills},
  author  = {Zheng, Boyuan and Fatemi, Michael Y. and Jin, Xiaolong and Wang, Zora Zhiruo and Gandhi, Apurva and Song, Yueqi and Gu, Yu and Srinivasa, Jayanth and Liu, Gaowen and Neubig, Graham and Su, Yu},
  journal = {arXiv preprint arXiv:2504.07079},
  year    = {2025}
}

@inproceedings{han2019learning,
  title     = {Learning to Discover Novel Visual Categories via Deep Transfer Clustering},
  author    = {Han, Kai and Vedaldi, Andrea and Zisserman, Andrew},
  booktitle = {Proceedings of the IEEE/CVF International Conference on Computer Vision},
  pages     = {8401--8409},
  year      = {2019}
}

@inproceedings{vaze2022generalized,
  title     = {Generalized Category Discovery},
  author    = {Vaze, Sagar and Han, Kai and Vedaldi, Andrea and Zisserman, Andrew},
  booktitle = {Proceedings of the IEEE/CVF Conference on Computer Vision and Pattern Recognition},
  pages     = {7492--7501},
  year      = {2022}
}

@inproceedings{zhang2022grow,
  title     = {Grow and Merge: A Unified Framework for Continuous Categories Discovery},
  author    = {Zhang, Xinwei and Jiang, Jianwen and Feng, Yutong and Wu, Zhi-Fan and Zhao, Xibin and Wan, Hai and Tang, Mingqian and Jin, Rong and Gao, Yue},
  booktitle = {Advances in Neural Information Processing Systems},
  volume    = {35},
  pages     = {27455--27468},
  year      = {2022}
}

@inproceedings{wu2023metagcd,
  title     = {{MetaGCD}: Learning to Continually Learn in Generalized Category Discovery},
  author    = {Wu, Yanan and Chi, Zhixiang and Wang, Yang and Feng, Songhe},
  booktitle = {Proceedings of the IEEE/CVF International Conference on Computer Vision},
  pages     = {1655--1665},
  year      = {2023}
}

@inproceedings{ma2024happy,
  title     = {Happy: A Debiased Learning Framework for Continual Generalized Category Discovery},
  author    = {Ma, Shijie and Zhu, Fei and Zhong, Zhun and Liu, Wenzhuo and Zhang, Xu-Yao and Liu, Cheng-Lin},
  booktitle = {Advances in Neural Information Processing Systems},
  volume    = {37},
  pages     = {50850--50875},
  year      = {2024}
}

@inproceedings{agarwal2024policy,
  title     = {On-Policy Distillation of Language Models: Learning from Self-Generated Mistakes},
  author    = {Agarwal, Rishabh and Vieillard, Nino and Zhou, Yongchao and Stanczyk, Piotr and Ramos Garea, Sabela and Geist, Matthieu and Bachem, Olivier},
  booktitle = {International Conference on Learning Representations},
  pages     = {21246--21263},
  year      = {2024}
}

@inproceedings{gu2024minillm,
  title     = {{MiniLLM}: Knowledge Distillation of Large Language Models},
  author    = {Gu, Yuxian and Dong, Li and Wei, Furu and Huang, Minlie},
  booktitle = {International Conference on Learning Representations},
  pages     = {32694--32717},
  year      = {2024}
}

@inproceedings{
wang2026egomemreason,
title={EgoMemReason: A Memory-Driven Reasoning Benchmark for Long-Horizon Egocentric Video Understanding},
author={Ziyang Wang and Yue Zhang and Shoubin Yu and Ce Zhang and Zengqi Zhao and Jaehong Yoon and Hyunji Lee and Gedas Bertasius and Mohit Bansal},
booktitle={Third Conference on Language Modeling},
year={2026},
url={https://openreview.net/forum?id=yD3YxhN6rN}
}

@InProceedings{kim2025uniskill,
  title = 	 {UniSkill: Imitating Human Videos via Cross-Embodiment Skill Representations},
  author =       {Kim, Hanjung and Kang, Jaehyun and Kang, Hyolim and Cho, Meedeum and Kim, Seon Joo and Lee, Youngwoon},
  booktitle = 	 {Proceedings of The 9th Conference on Robot Learning},
  pages = 	 {4269--4294},
  year = 	 {2025},
  publisher =    {PMLR},
}

@article{
wang2026from,
title={From Models to Systems: A Comprehensive Survey of Efficient Multimodal Learning},
author={Pan Wang and Siwei Song and Hui Ji and Siqi Cao and Heng Yu and Zhijian Liu and Huanrui Yang and Yingyan Celine Lin and Beidi Chen and Mohit Bansal and Xiaoming Liu and Pengfei Zhou and Ming-Hsuan Yang and Tianlong Chen and Jingtong Hu},
journal={Transactions on Machine Learning Research},
issn={2835-8856},
year={2026},
url={https://openreview.net/forum?id=yfTU8FTS2Z}
}

@inproceedings{zhang2026lens,
  title={LENS: Adaptive Spatio-Temporal Zooming for Keyframe Sampling in Long-Form Videos},
  author={Zhang, Ce and He, Jinxi and Sycara, Katia and Xie, Yaqi},
  booktitle={European Conference on Computer Vision},
  pages={511--529},
  year={2026},
  organization={Springer}
}

@inproceedings{
luo2026pyspatial,
title={pySpatial: Generating 3D Visual Programs for Zero-Shot Spatial Reasoning},
author={Zhanpeng Luo and Ce Zhang and Silong Yong and Cunxi Dai and Qianwei Wang and Haoxi Ran and Guanya Shi and Katia P. Sycara and Yaqi Xie},
booktitle={The Fourteenth International Conference on Learning Representations},
year={2026},
url={https://openreview.net/forum?id=yv15C8ql24}
}

@InProceedings{zhang2026evolving,
    author    = {Zhang, Ce and He, Jinxi and He, Junyi and Sycara, Katia and Xie, Yaqi},
    title     = {Evolving Contextual Safety in Multi-Modal Large Language Models via Inference-Time Self-Reflective Memory},
    booktitle = {Proceedings of the IEEE/CVF Conference on Computer Vision and Pattern Recognition},
    month     = {June},
    year      = {2026},
    pages     = {41182-41192}
}

@article{
zhang2026vscan,
title={{VS}can: Rethinking Visual Token Reduction for Efficient Large Vision-Language Models},
author={Ce Zhang and Kaixin Ma and Tianqing Fang and Wenhao Yu and Hongming Zhang and Zhisong Zhang and Haitao Mi and Dong Yu},
journal={Transactions on Machine Learning Research},
issn={2835-8856},
year={2026},
url={https://openreview.net/forum?id=KZYhyilFnt},
note={}
}

@article{zhang2026streamscout,
  title   = {{StreamScout}: Learning When to Look Deeper for Streaming Video Understanding},
  author  = {Zhang, Ce and Bi, Jing and He, Jinxi and Zhang, Jianshu and Lin, Jingyang and Xiao, Yunzhong and Fu, Minghao and Xie, Yaqi and Xie, Zhentao and Chen, Weicong and Sycara, Katia and Zhou, Ming},
  journal = {arXiv preprint arXiv:2609.00291},
  year    = {2026}
}

@inproceedings{pi2024image,
  title     = {Image Textualization: An Automatic Framework for Creating Accurate and Detailed Image Descriptions},
  author    = {Pi, Renjie and Zhang, Jianshu and Zhang, Jipeng and Pan, Rui and Chen, Zhekai and Zhang, Tong},
  booktitle = {Advances in Neural Information Processing Systems},
  year      = {2024}
}

@inproceedings{zhang2026progresslm,
  title     = {{ProgressLM}: Towards Progress Reasoning in Vision-Language Models},
  author    = {Zhang, Jianshu and Qian, Chengxuan and Sun, Haosen and Lu, Haoran and Wang, Dingcheng and Xue, Letian and Liu, Han},
  booktitle = {Proceedings of the Annual Meeting of the Association for Computational Linguistics},
  year      = {2026}
}

@inproceedings{zhang2025vlm2,
  title     = {{VLM$^2$-Bench}: A Closer Look at How Well {VLMs} Implicitly Link Explicit Matching Visual Cues},
  author    = {Zhang, Jianshu and Yao, Dongyu and Pi, Renjie and Liang, Paul Pu and Fung, Yi R.},
  booktitle = {Proceedings of the Annual Meeting of the Association for Computational Linguistics},
  year      = {2025}
}

@inproceedings{zhang2024core,
  title     = {{CORE}: Mitigating Catastrophic Forgetting in Continual Learning through Cognitive Replay},
  author    = {Zhang, Jianshu and Fu, Yankai and Peng, Ziheng and Yao, Dongyu and He, Kun},
  booktitle = {Proceedings of the Annual Meeting of the Cognitive Science Society},
  year      = {2024}
}

@inproceedings{rong2025can,
  title     = {{CAN}: Leveraging Clients as Navigators for Generative Replay in Federated Continual Learning},
  author    = {Rong, Xuankun and Zhang, Jianshu and He, Kun and Ye, Mang},
  booktitle = {International Conference on Machine Learning},
  year      = {2025}
}

@inproceedings{wang2026explore,
  title={WebAggregator: Enhancing Compositional Reasoning Capabilities of Deep Research Agent Foundation Models},
  author={Wang, Rui and Zhang, Ce and Ma, Jun-Yu and Zhang, Jianshu and Wang, Hongru and Chen, Yi and Xue, Boyang and Fang, Tianqing and Zhang, Zhisong and Zhang, Hongming and others},
  booktitle={Proceedings of the 64th Annual Meeting of the Association for Computational Linguistics},
  pages={24486--24517},
  year={2026}
}

@article{zhang2026progress,
  title   = {Progress Reward Modeling for Robotic Learning: A Comprehensive Survey},
  author  = {Zhang, Jianshu and Wu, Keliang and Lu, Haoran and Liu, Anbang and Zhang, Ce and Yin, Weijie and Qian, Chengxuan and Yang, Xiyuan and Pan, Zhenyu and Ye, Guo and Liu, Han},
  journal = {arXiv preprint arXiv:2607.21655},
  year    = {2026}
}

@inproceedings{pi2025personalized,
  title     = {Personalized Visual Instruction Tuning},
  author    = {Pi, Renjie and Zhang, Jianshu and Han, Tianyang and Zhang, Jipeng and Pan, Rui and Zhang, Tong},
  booktitle = {International Conference on Learning Representations},
  year      = {2025}
}

@inproceedings{brown2020language,
  title     = {Language Models are Few-Shot Learners},
  author    = {Brown, Tom and Mann, Benjamin and Ryder, Nick and Subbiah, Melanie and Kaplan, Jared D. and Dhariwal, Prafulla and Neelakantan, Arvind and Shyam, Pranav and Sastry, Girish and Askell, Amanda and Agarwal, Sandhini and Herbert-Voss, Ariel and Krueger, Gretchen and Henighan, Tom and Child, Rewon and Ramesh, Aditya and Ziegler, Daniel and Wu, Jeffrey and Winter, Clemens and Hesse, Chris and Chen, Mark and Sigler, Eric and Litwin, Mateusz and Gray, Scott and Chess, Benjamin and Clark, Jack and Berner, Christopher and McCandlish, Sam and Radford, Alec and Sutskever, Ilya and Amodei, Dario},
  booktitle = {Advances in Neural Information Processing Systems},
  pages     = {1877--1901},
  year      = {2020}
}

@article{ning2026code,
  title   = {Code as Agent Harness},
  author  = {Ning, Xuying and Tieu, Katherine and Fu, Dongqi and Wei, Tianxin and Li, Zihao and Bei, Yuanchen and Zou, Jiaru and Ai, Mengting and Liu, Zhining and Li, Ting-Wei and others},
  journal = {arXiv preprint arXiv:2605.18747},
  year    = {2026}
}

@article{baker2009action,
  title   = {Action Understanding as Inverse Planning},
  author  = {Baker, Chris L. and Saxe, Rebecca and Tenenbaum, Joshua B.},
  journal = {Cognition},
  volume  = {113},
  number  = {3},
  pages   = {329--349},
  year    = {2009}
}
\bibliographystyle{iclr2027_conference}

\end{document}